\documentclass[a4paper,fleqn]{cas-dc}

\usepackage[authoryear]{natbib}
\hypersetup{
    colorlinks=true,
    linkcolor=blue,
    citecolor=blue,
    urlcolor=blue,
    bookmarks=false
}

\usepackage{graphicx}
\usepackage{subcaption}
\usepackage{float}
\usepackage{placeins} 
\usepackage{tabularx}
\usepackage{array}

\usepackage{algorithm}
\usepackage{algpseudocode}
\usepackage{amsmath}
\usepackage{amssymb}
\usepackage{makecell}

\usepackage{enumitem}

\def\tsc#1{\csdef{#1}{\textsc{\lowercase{#1}}\xspace}}
\tsc{WGM}
\tsc{QE}
\tsc{EP}
\tsc{PMS}
\tsc{BEC}
\tsc{DE}

\begin{document}
\let\WriteBookmarks\relax
\def\floatpagepagefraction{1}
\def\textpagefraction{.001}

\ExplSyntaxOn
\RenewDocumentCommand \printorcid {} {}
\ExplSyntaxOff

\shorttitle{}    
\shortauthors{}  

\title [mode = title]{Comparison-Based Ordinal Learning for Proactive Driving Risk Assessment}

\author[1]{Zhuoren Li}
\fnmark[1]
\author[1]{Yi Zhong}
\fnmark[1]
\author[1]{Weiqi Zhang}
\author[1]{Xinrui Zhang}
\author[1]{Lu Xiong}
\author[2]{Chongfeng Wei}
\author[1]{Bo Leng}
\cormark[1]
\ead{lengbo@tongji.edu.cn}

\affiliation[1]{
    organization={College of Automotive and Energy Engineering, Tongji University},
    city={Shanghai},
    country={China}
}
\affiliation[2]{
    organization={James Watt School of Engineering, University of Glasgow},
    city={Glasgow},
    country={UK}
}

\cortext[1]{Corresponding author}
\fntext[1]{These authors contributed equally to this work.}

\begin{abstract}
Real-time driving risk assessment provides an essential basis for proactive safety by identifying and quantifying the danger of ongoing road interactions before adverse outcomes occur. In automated vehicles, such risk assessment can be further embedded into the decision-making and planning process to guide risk-aware maneuver generation.
However, due to the scarcity of collision data and frame-level risk labels, existing driving risk assessment methods often rely on surrogate objectives, which may imperfectly align with true collision risk and not faithfully reflect the relative danger of driving interaction. This paper proposes a comparison-based ordinal risk learning framework that learns collision-relevant risk scores from pairwise supervision in driving data, directly modeling relative risk ordering without requiring numerical frame-level risk labels. We derive pairwise comparisons from three sources of event-structured driving data for such ordinal risk learning: temporal progression within safety-critical sequences, event-level contrast between dangerous and normal interactions, and physics-based counterfactual perturbations. On this basis, instantiations with three risk-scoring function parameterizations are implemented, including directly learning risk scores from comparison data, and aligning existing single or multiple surrogate-based risk models. The proposed framework is evaluated on the 100-Car and SHRP2 naturalistic driving datasets using a proactive collision warning task. Results show that the proposed framework improves high-recall risk discrimination, warning precision, and warning lead time over representative surrogate-based baselines across both in-distribution and out-of-distribution evaluations. These results suggest that the proposed framework can contribute to proactive safety research by providing more reliable risk assessment for automated driving systems and safety-critical driving interactions.
\end{abstract}



\maketitle

\section{Introduction}
\label{sec:intro}

Road traffic crashes remain a major global public health challenge, causing approximately 1.19 million deaths annually worldwide \citep{WHO2023}. Unlike reactive safety analysis, which primarily explains crashes after they occur, proactive driving risk assessment aims to identify potentially dangerous traffic interactions in real time and support early intervention before they escalate into collisions \citep{SSMreview}. These capabilities are essential for a wide range of applications, for example Advanced Driver-Assistance Systems (ADAS), high-level autonomous driving systems, infrastructure design and operations \citep{Jiao2026GSSM, JinTNNLS}.

The fundamental challenge is that collision risk is a latent, future-oriented quantity rather than a directly observable frame-level label.
However, collision events are rare in naturalistic driving; even large-scale studies such as the 100-Car Naturalistic Driving Study (NDS) \citep{Dingus2006} contain only a limited number of crashes and near-crashes among millions of recorded miles. Consequently, assigning a reliable numerical risk score to every driving moment would require dense, fine-grained supervision that is generally unavailable in naturalistic datasets.

These difficulties have led much of the research to rely on surrogate objectives that estimate collision risk indirectly. Existing approaches include analytical kinematic proxies derived from relative motion \citep{Guo2023TTC2D, Laureshyn2010TAdv, Cheng2025EI}, forward-prediction approaches that transform predicted future motion or occupancy into prospective conflict-risk estimates~\citep{Yao2025ICRA, Liu2026RiskNet, LiTITS}, and emerging distribution-modeling approaches that learn normal interaction distributions \citep{jiao2025unified, Jiao2026GSSM}.
Despite their methodological differences, these approaches provide practical alternatives when large-scale collision annotations are unavailable.

However, these approaches largely depend on predefined surrogate assumptions or independently constructed intermediate models. Kinematic indicators depend on simplified motion extrapolations, which may not fully capture how risk develops when road users deviate from these assumptions; Forward-prediction approaches infer risk from separately developed trajectory predictors through predicted conflict, collision probability, or potential-field-based risk measures, making the final risk estimate sensitive to both prediction errors and the subsequent risk-mapping design; Distribution-modeling approaches assess risk through identifying statistically rare interaction patterns from normal driving behaviour distribution, but statistically rare states are not necessarily safety-critical, and frequently observed states can still become hazardous under specific interaction conditions. Although these formulations are useful and interpretable within their respective assumptions, their connection to the observed spatiotemporal development and terminal outcomes of crash and near-crash events remains indirect. This may limit their ability to consistently characterize when an interaction is becoming dangerous, leading to false alarms, missed hazardous interactions, or delayed warning triggers in proactive safety applications.

We argue that, despite the scarcity of collision or near-collision events in naturalistic driving data, collision-relevant supervision need not rely solely on point-wise numerical risk labels. These safety-critical events contain more information than a single event-level binary outcome. Each event contains structured evidence of how an ordinary interaction evolves toward a safety-critical endpoint, while conflict-free interactions and basic physical constraints provide additional reference information about lower-risk conditions. Such event-structured information offers a weak but meaningful form of supervision for learning driving risk, without requiring an explicit numerical risk value for every frame. From this perspective, the key challenge is to transform limited crash-relevant data into learnable ordinal constraints that make fuller use of scarce safety-critical events.

Building on this idea, this paper proposes a comparison-based ordinal learning framework for proactive driving risk assessment. Instead of requiring calibrated frame-level risk labels, the proposed framework converts event-structured information from crash and near-crash records, conflict-free event contrast, and physics-based counterfactual perturbations into ordinal supervision over driving observations. This supervision is formulated as a pairwise ordinal learning objective for a scalar risk scoring function, and the proposed framework itself is not tied to a specific risk scoring architecture. To the best of our knowledge, this is the first study to formulate proactive driving risk assessment as comparison-based ordinal risk-score learning problem. Beyond learning a risk score from scratch, the proposed framework further enables ordinal alignment of existing surrogate-based risk scores and their integration into a unified risk representation. The main contributions of this paper include:

\begin{itemize}[leftmargin=*, nosep]
\item We propose a comparison-based ordinal learning framework for proactive driving risk assessment, which learns a collision-relevant risk score without requiring calibrated frame-level numerical risk labels.

\item We construct multi-source pairwise comparison supervision from scarce safety-critical events and related interaction data, covering temporal progression, event contrast, and physics-based counterfactual evidence, and learn from these comparisons through source-weighted ranking and collision-endpoint anchoring.

\item We instantiate the framework through three risk-scoring function parameterizations: direct neural risk-score learning, single-surrogate and multi-surrogate risk alignment.

\end{itemize}

The remainder of this paper is organised as follows. Section~\ref{sec:related} reviews related work. Section~\ref{sec:method} details the proposed risk learning framework and its three instantiations. Section~\ref{sec:experiments} describes the experimental setup and reports results. Section~\ref{sec:conclusion} finally concludes the paper.

\section{Related Work}
\label{sec:related}

Estimating collision risk in real time is a long-standing problem in traffic safety research. Collision events are rare in naturalistic driving, and assigning reliable frame-level risk labels to individual driving observations is infeasible at current data collection scales. The literature deals with this constraint along two broad lines. One line avoids collision supervision altogether, relying on surrogate objectives whose outputs approximate risk through physically or statistically motivated proxies. The other line uses collision-level labels directly but is constrained by their granularity and volume. In addition, ordinal and pairwise learning has been widely studied as a weak-supervision paradigm when calibrated numerical labels are difficult to obtain.

\subsection{Surrogate-Based Risk Quantification}

Surrogate-based risk quantification avoids direct supervision from collision outcomes by defining risk through observable proxy information. These proxies vary in their grounding, including kinematic indicators, forward-prediction-based estimates, and distribution-based anomaly scores. Despite their practical value, their outputs are primarily determined by surrogate assumptions or intermediate modeling criteria, rather than by direct calibration against observed collision outcomes.

\subsubsection{Kinematic Safety Measures}

Kinematic surrogate safety measures constitute one of the earliest and most widely used families of risk indicators. They estimate collision risk from instantaneous relative motion between road users under simplified short-horizon motion assumptions. Time-to-Collision (TTC) \citep{Hayward1972NearMiss} measures the remaining time until geometric overlap under constant-velocity extrapolation. Extensions such as TTC2D \citep{Guo2023TTC2D} generalize this idea to two-dimensional motion, thereby capturing lateral components of potential conflict. Time Advantage (TAdv) \citep{Laureshyn2010TAdv} measures the temporal margin between two road users passing a common conflict point, while Emergency Index (EI) \citep{Cheng2025EI} quantifies the evasive deceleration required to avoid encroachment. These measures are computationally inexpensive, require no training data, and provide clear physical interpretations.

Their main limitation is that risk is inferred from predefined kinematic assumptions. Constant-motion extrapolation, prescribed evasive-response models, and fixed conflict-point definitions can be effective in simple interactions, but they may become unreliable when agents accelerate, steer, react strategically, or interact with multiple surrounding vehicles. As a result, the resulting risk scores may not consistently reflect how collision risk evolves under complex and uncertain driving interactions.

\subsubsection{Forward-Prediction-Based Risk Scoring}

To obtain a more flexible characterization of future interactions, some work estimates risk by first predicting future trajectories states and then mapping these predictions into prospective collision-risk scores. Some methods compute probabilistic risk metrics from multi-modal trajectory predictions, thereby linking future motion uncertainty to collision likelihood \citep{Probrisk}. Other studies couple data-driven trajectory prediction with spatial risk fields or learned risk representations to evaluate future interaction hazards \citep{Yao2025ICRA, Liu2026RiskNet, Wang2025DynamicRisk}. Occupancy-based methods further convert predicted future states into probabilistic occupancy maps or spatio-temporal occupancy heatmaps, and assess risk through occupancy overlap, collision checking, or safety verification along the ego vehicle's intended trajectory \citep{Wang2025DynamicRisk, zheng2026preoccmap}. Compared with kinematic rules, these methods can represent non-constant-velocity motion, account for uncertainty in future behavior, and provide a more flexible basis for evaluating potential conflicts.

A recurring difficulty in this paradigm is that prediction and risk scoring are often optimized or designed according to different criteria. Prediction modules are commonly trained to reduce average trajectory error over the data distribution, whereas proactive risk assessment is most sensitive to rare collision-critical outcomes and tail interactions. The subsequent risk-mapping component, whether based on probabilistic collision calculation, occupancy overlap, safety verification, or a risk field, is typically specified separately from the predictor. As a result, improvements in average prediction accuracy do not necessarily translate into more reliable risk estimates, and a risk-mapping rule that works with one predictor may not remain appropriate when the predictor is retrained or replaced. Thus, although prediction-based approaches provide a richer description of future interactions than kinematic SSMs, their risk scores remain indirectly tied to observed collision outcomes.

\subsubsection{Distribution-Based Anomaly Scoring}

A further family of surrogate methods grounds risk in statistical deviation from normal driving behavior. The core assumption is that dangerous interactions are statistically rare relative to the normal traffic distribution. Under this view, observations located in safety-relevant tails or low-likelihood regions of a learned interaction distribution can be assigned higher risk scores.

The Generalized Surrogate Safety Measure (GSSM) \citep{Jiao2026GSSM} is representative of this approach. GSSM trains a conditional model on large-scale naturalistic data to
learn the distribution of inter-vehicle spacing given interaction context, and scores each observation as a percentile within the lower tail of this distribution. Unusually tight spacing relative to the prevailing context is flagged as risky, allowing the
model to adapt its signal to heterogeneous traffic conditions. REDOUBT \citep{Wang2025REDOUBT} adopts a complementary strategy, using flow matching to model the in-distribution density of a learned latent space; this formulation accommodates more complex input distributions and assigns elevated risk to observations with low likelihood under the learned flow. Lyu et al. \citep{lyu2026unsupervised} model normal multi-agent driving distributions with a Transformer-based trajectory predictor and identify anomalous scenarios through prediction residuals, while further evaluating their physical alignment with established SSMs.

These methods scale naturally with data volume and require no crash outcome labels. However, statistical unusualness and collision risk are not always equivalent. An infrequent but physically safe configuration may be assigned a high risk score because it is unusual under the learned representation, whereas a frequently observed interaction pattern may still be hazardous under specific conditions that are not fully captured by the distributional model. Thus, distribution-based methods provide an adaptive data-driven surrogate for risk, but their scores remain tied to statistical extremeness rather than directly to observed collision outcomes.


Across all three families, surrogate objectives provide no direct link to collision outcomes. Their scores are therefore shaped by kinematic assumptions, prediction objectives, risk-mapping rules, or distributional normality, rather than being directly constrained by observed collision outcomes or the temporal progression of safety-critical events. This indirect grounding may lead to false alarms when an unusual but safe state receives a high surrogate score, missed warnings when a genuinely hazardous configuration produces only moderate surrogate values, and score dynamics that diverge from actual risk accumulation as a conflict develops.

\subsection{Supervised Methods for Accident Anticipation}

Early studies formulate accident anticipation as sequential prediction from dashcam videos, using recurrent or attention-based models to identify visual and object-level cues that precede an accident \citep{Chan2017TAA}. Subsequent methods improve this paradigm by modeling spatio-temporal relations among traffic agents and incorporating prediction uncertainty \citep{Bao2020TAA}, or by detecting abnormal events through future-location prediction and residual-based inconsistency \citep{Yao2019TAA}. More recent work further reformulates accident anticipation as temporal occurrence prediction, aiming to predict when an accident will occur rather than only assigning frame-level anomaly scores \citep{Zhao2025TOP}. Recently, RiskProp \citep{Zou_2026_CVPR_RiskProp} leverages reliably annotated collision frames as anchors and propagates risk signals backward through self-supervised temporal constraints, producing smoother risk curves before collisions. Compared with surrogate-based risk measures, these studies are more closely connected to accident outcomes and early-warning objectives.

In practice, however, two challenges arise. First, genuine collision events are extremely scarce in naturalistic driving: even large-scale studies such as the 100-Car NDS \citep{Dingus2006} contain only hundreds of crashes across millions of miles, making dense frame-level supervision infeasible. Second, event-level binary labels are shared by every frame in a crash sequence and carry no information about when within the sequence risk escalates; frames recorded seconds before impact receive the same label as frames at the moment of impact.

Some works address data scarcity through simulation. Schoonbeek et al.\citep{Schoonbeek2022IV} pre-train a feature extractor on real data and fine-tune on simulated crash events; Lu et al. \citep{Lu2025PhysicsInformed} use a physics-based model to augment real records with synthetic conflict scenarios. Even with simulated data, these methods depend on per-sample labels whose reliable annotation remains difficult in real-world settings and whose distributional properties in simulation diverge from naturalistic driving.

\subsection{Learning from Pairwise Comparisons}

Pairwise comparison learning has a long history in settings where calibrated absolute scores are difficult to obtain but relative orderings are available. In information retrieval, Joachims \citep{Joachims2002} showed that click-through data can implicitly encode pairwise relevance preferences for learning ranking models without explicit relevance labels. In reinforcement learning from human feedback, pairwise preference judgments are widely used to train reward models when direct reward specification is difficult \citep{Christiano2017}. The Bradley--Terry model \citep{Bradley1952}, which underlies many pairwise preference-learning approaches, provides a probabilistic framework for translating noisy comparisons into a scalar ranking function.

This idea is particularly relevant to proactive driving risk assessment. Judging which of two traffic situations is more safety-critical is often more tractable than assigning a calibrated collision probability to each individual observation. More importantly, ordinal supervision can be derived from event-structured naturalistic driving data without per-frame risk labels. In this sense, pairwise comparison learning provides a bridge between surrogate-based risk scoring and fully supervised accident anticipation: it reduces the reliance on manually specified proxy objectives while avoiding the need for dense numerical risk labels, thereby retaining the flexibility of data-driven learning under limited safety-critical data. Despite this relevance, to the best of our knowledge, comparison-based ordinal learning has not been systematically explored for learning proactive driving risk scores from safety-critical event structures. This work develops such a framework to learn collision-relevant risk-scoring functions without requiring calibrated frame-level numerical risk labels.

\begin{figure*}
  \centering
  \includegraphics[width=\linewidth]{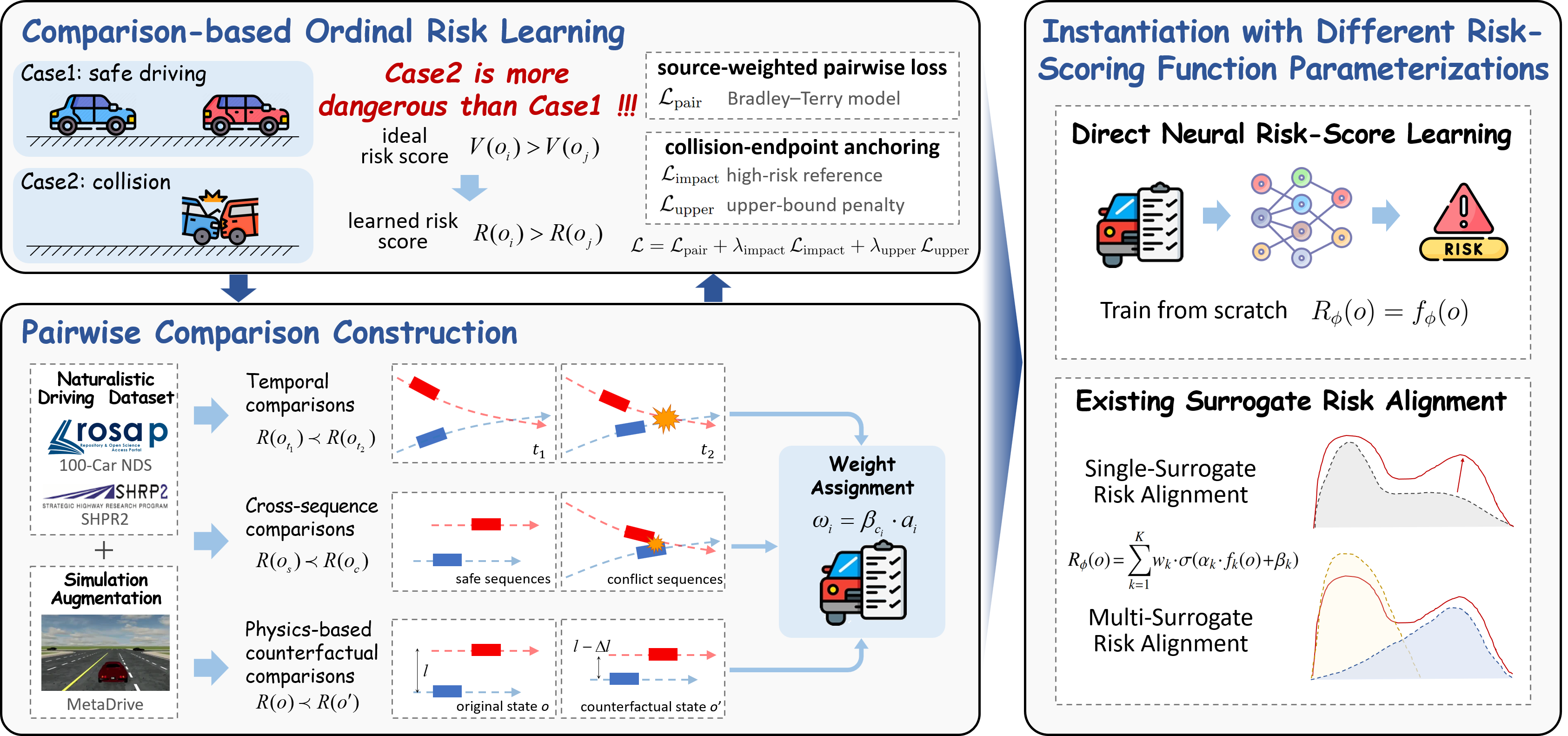}
  \caption{The overall framework of our proposed comparison-based ordinal risk learning.}
  \label{fig:framework}
\end{figure*}

\section{Method}
\label{sec:method}

Figure~\ref{fig:framework} provides an overview of the proposed comparison-based ordinal risk learning framework. The method consists of three components: (i) construction of pairwise comparisons from naturalistic and simulated driving data, (ii) ordinal risk learning via a Bradley–Terry objective with collision-endpoint anchoring, and (iii) instantiation with different risk-scoring function parameterizations.

\subsection{Problem Formulation}
\label{sec:problem}

Proactive driving risk assessment assigns to each driving observation a continuous driving risk score reflecting how likely the current interaction is to escalate into a collision. Let $o$ denote the driving observation capturing the interaction state (e.g., vehicle kinematics, spacing, and road context). The ideal scoring target is the conditional probability:
\begin{equation}
V(o) = P\!\left(C = 1 \mid o\right),
\end{equation}
where $C \in \{0,1\}$ indicates whether a collision occurs in the near future given the current observation, under the assumption of locally stationary interaction dynamics over a short time horizon.

Recovering $V(o)$ numerically requires dense, frame-level supervision that naturalistic driving data cannot provide, since collision events ($C=1$) are extremely sparse in real-world driving logs.

For proactive collision-risk warning, exact numerical recovery of $V$ is not the essential requirement. A reliable warning system depends on the learned risk scoring function $R_\phi$ preserving the correct risk ordering: whenever $o_i$ is more dangerous than $o_j$,
\begin{equation}
V(o_i) > V(o_j) \implies R_\phi(o_i) > R_\phi(o_j).
\label{eq:rank_objective}
\end{equation}
This ordinal objective is strictly weaker than numerical recovery, yet sufficient for risk-ordered warning and decision-making. We operationalise this through pairwise comparisons. Given a pair $(o_w, o_l)$ where $o_w$ is judged more risky, the Bradley--Terry (BT) model \citep{Bradley1952} assigns the ordering probability:
\begin{equation}
P(o_w \succ o_l)=
\frac{e^{R_\phi(o_w)}}
{e^{R_\phi(o_w)}+e^{R_\phi(o_l)}}
=\sigma\!\left(R_\phi(o_w)-R_\phi(o_l)\right),
\end{equation}
with negative log-likelihood:
\begin{equation}
\ell_{\mathrm{BT}}(o_w, o_l) =
-\log\sigma\!\left(R_\phi(o_w) - R_\phi(o_l)\right).
\end{equation}
The BT model tolerates noisy individual comparisons: the correct ordering becomes increasingly certain as the true risk gap grows. The key challenge is therefore to construct reliable pairwise comparisons from scarce safety-critical events, each documenting how an ordinary interaction escalates toward a collision, and to use them to train $R_\phi$ toward condition~(\ref{eq:rank_objective}).

The rest of this section describes how comparisons are constructed from event-structured data, how they form the training objective, and how the framework is instantiated under three parameterizations.

\subsection{Pairwise Comparison Construction}
\label{sec:comparisons}

Translating event-structured driving data into learnable supervision means
extracting pairs of observations for which the ordering of risk can be
asserted.
We formalise this as a dataset of pairwise comparisons
$\mathcal{D} = \{(o_w^{(i)}, o_l^{(i)})\}_{i=1}^{N}$,
where $o_w^{(i)} \succ o_l^{(i)}$ denotes that observation $o_w^{(i)}$ is
judged more dangerous than $o_l^{(i)}$.
We construct three comparison sources, each exploiting a
different structural property of driving data.

\paragraph{Temporal comparisons.}
Within a single conflict sequence that culminates in a crash or near-crash,
risk generally increases as the moment of impact approaches.
We exploit this temporal structure by sampling pairs $(o_{t_2}, o_{t_1})$
from the same event with $t_1 < t_2 \le t_{\mathrm{impact}}$ and
asserting $o_{t_2} \succ o_{t_1}$.
To limit the noise introduced by comparisons spanning long temporal gaps,
we restrict the maximum separation between paired frames to a bounded interval.
These comparisons provide fine-grained, within-event supervision that
captures the dynamic escalation of risk as a conflict develops toward its endpoint.

\paragraph{Cross-sequence comparisons.}
At the event level, conflict sequences are drawn from a fundamentally
riskier regime than normal, conflict-free driving sequences.
For an observation $o_c$ sampled from a conflict event and an observation
$o_{nc}$ sampled from a normal driving sequence, we assert $o_c \succ o_{nc}$.
This yields a coarser but broader supervision signal: rather than characterising
within-event risk dynamics, it calibrates the global risk separation between
dangerous and benign interaction regimes.
To prevent long sequences from dominating the comparison pool, pairs are
constructed by first drawing uniformly over events and then uniformly over
time within each selected event.

\paragraph{Physics-based counterfactual comparisons.}
Physical reasoning provides a third source of supervision that is entirely independent of event outcomes. Under locally comparable interaction conditions, several risk-relevant variables have directionally interpretable effects on collision risk. For example, decreasing inter-vehicle headway, increasing closing speed, reducing lateral clearance, and increasing vehicle dimensions is expected to make the interaction more safety-critical. Based on this intuition, we construct these comparisons by taking a baseline observation $o$ and applying a risk-increasing perturbation to a single physical variable to obtain $o'$, asserting $o' \succ o$.
Because perturbations can be applied to both conflict and normal observations without knowledge of event outcomes, this source provides broad coverage of the interaction state space.
Physically-grounded comparisons act as local shape constraints: they ensure that the scoring function responds in the correct direction to risk-relevant variables, independent of what the event-level data supply.

\paragraph{Simulation-based comparison augmentation.}
To expand the comparison pool at low cost, we additionally generate conflict and normal interaction sequences in MetaDrive \citep{Li2022} and apply the same three construction procedures to simulated data. Because simulated interactions differ distributionally from real-world driving, we treat simulation data as a supplementary source rather than a primary one,
and evaluate its contribution empirically in Section~\ref{sec:ablation}.

All comparison types are constructed exclusively within the training partition to prevent label leakage; the full construction pipeline is summarized in Algorithm~\ref{alg:comparison}.

\begin{algorithm}[t]
\small
\caption{Construction of Pairwise Comparison Data from Naturalistic Driving}
\label{alg:comparison}
\begin{algorithmic}[1]
\Require Conflict sequences $\mathcal{S}_c=\{(\tau_e,t_{\mathrm{impact},e})\}$;
safe sequences $\mathcal{S}_{nc}$;
temporal gap bounds $[\Delta_{\min},\Delta_{\max}]$;
risk-increasing variables $\mathcal{V}$;
target number of pairs per source $M$
\Ensure Pairwise comparison dataset
$\mathcal{D}=\mathcal{D}_{\mathrm{temp}}\cup
\mathcal{D}_{\mathrm{cross}}\cup
\mathcal{D}_{\mathrm{phys}}$

\Statex
\State \textbf{Step 1: Feature extraction}
\For{each sequence $\tau \in \mathcal{S}_c \cup \mathcal{S}_{nc}$}
    \State Compute observation $o_t$ for each frame $t \in \tau$
\EndFor

\Statex
\State \textbf{Step 2: Temporal comparisons}
\State $\mathcal{D}_{\mathrm{temp}} \gets \emptyset$
\For{each conflict sequence $(\tau_e,t_{\mathrm{impact},e}) \in \mathcal{S}_c$}
    \For{$m = 1$ to $M / |\mathcal{S}_c|$}
        \State Sample $(t_1,t_2)$ such that
        $t_1 < t_2 \le t_{\mathrm{impact},e}$ and
        $\Delta_{\min} \le t_2-t_1 \le \Delta_{\max}$
        \State Add $(o_{t_2}, o_{t_1})$ to $\mathcal{D}_{\mathrm{temp}}$
        \Comment{later frame is riskier}
    \EndFor
\EndFor

\Statex
\State \textbf{Step 3: Cross-sequence comparisons}
\State $\mathcal{D}_{\mathrm{cross}} \gets \emptyset$
\For{$m = 1$ to $M$}
    \State Sample conflict sequence $(\tau_e,t_{\mathrm{impact},e}) \sim \mathcal{S}_c$
    \State Sample frame $t_c$ from $\tau_e$
    \State Sample safe sequence $\tau' \sim \mathcal{S}_{nc}$
    \State Sample frame $t_{nc}$ from $\tau'$
    \State Add $(o_{t_c}, o_{t_{nc}})$ to $\mathcal{D}_{\mathrm{cross}}$
    \Comment{conflict frame is riskier}
\EndFor

\Statex
\State \textbf{Step 4: Physics-based counterfactual comparisons}
\State $\mathcal{D}_{\mathrm{phys}} \gets \emptyset$
\For{each sequence $\tau \in \mathcal{S}_c \cup \mathcal{S}_{nc}$}
    \For{each frame $t \in \tau$}
        \For{each variable $v \in \mathcal{V}$}
            \State Construct $o'_t$ by applying a risk-increasing perturbation to $v$ in $o_t$
            \State Add $(o'_t,o_t)$ to $\mathcal{D}_{\mathrm{phys}}$
            \Comment{perturbed frame is riskier}
        \EndFor
    \EndFor
\EndFor
\State Randomly subsample $\mathcal{D}_{\mathrm{phys}}$ to $M$ pairs

\Statex
\State \Return $\mathcal{D}_{\mathrm{temp}}\cup
\mathcal{D}_{\mathrm{cross}}\cup
\mathcal{D}_{\mathrm{phys}}$
\end{algorithmic}
\end{algorithm}

\subsection{Comparison-Based Ordinal Learning Objective}
\label{sec:objective}

Given the pairwise comparison dataset $\mathcal{D}$ constructed in
Section~\ref{sec:comparisons}, we now define the training objective for
$R_\phi$.
Applying the Bradley--Terry loss from Section~\ref{sec:problem} directly
across all comparisons would shape the ordinal structure of $R_\phi$,
but leave its absolute scale undetermined.
The objective consists of two complementary components. Source-weighted pairwise ranking aggregates multi-source comparison supervision with effectiveness-aware weights to train $R_\phi$ toward the correct risk ordering. Collision-endpoint anchoring regularises the absolute scale by tying impact-imminent states (representing the most dangerous and consistently interpretable condition observable in the data) to a fixed high-risk reference value, grounding the learned score in real collision outcomes.

\paragraph{Source-weighted pairwise ranking.}
The three comparison sources provide ordering supervision of different types and at different granularities, and individual comparisons within each source vary in how strongly they support the asserted ordering.
To reflect the varying reliability and evidence strength, we assign each comparison a composite weight $w_i = \beta_{c_i} \cdot a_i$, decomposed into a source-level coefficient $\beta_{c_i}$ and a within-source sample weight $a_i$, where $c_i \in \{\mathrm{cross},\,\mathrm{temporal},\,\mathrm{physics}\}$
denotes the source type of comparison $i$.

The source-level coefficients assign different overall importance to each
comparison type:
\begin{equation}
\beta_{c_i} =
\begin{cases}
\beta_{\mathrm{temporal}}, & c_i = \text{temporal}, \\
\beta_{\mathrm{cross}},    & c_i = \text{cross-sequence}, \\
\beta_{\mathrm{physics}},  & c_i = \text{physics-based counterfactual}.
\end{cases}
\end{equation}

Within each source, the sample weight $a_i$ further differentiates pairs
by the strength of their ordering evidence.
Cross-sequence pairs are treated uniformly ($a_i = 1$), as all
conflict--normal pairings represent equally valid global ordering constraints.
For temporal comparisons, stronger evidence corresponds to a larger time-to-impact gap between the two frames; we therefore weight each pair by the normalised positive time-to-impact difference:
\begin{equation}
a_i =
\frac{\max(\delta_{\mathrm{tti}, i},\, 0)}
     {\mathbb{E}_{\mathrm{temporal}}\!\left[\max(\delta_{\mathrm{tti}}, 0)\right] + \epsilon},
\qquad
c_i = \mathrm{temporal}.
\end{equation}
where $\delta_{\mathrm{tti},i} = \mathrm{TTI}(o_l^{(i)}) - \mathrm{TTI}(o_w^{(i)})$
is the time-to-impact difference between the less-risky and more-risky
observations in pair $i$, and $\epsilon > 0$ is a small constant for
numerical stability.
Normalization keeps the mean within-source weight at unity, preventing
this source from dominating due to scale. For physics-based comparisons, the analogous evidence measure is the normalized perturbation magnitude:
\begin{equation}
\begin{aligned}
&a_i =
\frac{\max(\delta_{\mathrm{phys}, i},\, 0)}
     {\mathbb{E}_{\mathrm{physics}}\!\left[\max(\delta_{\mathrm{phys}}, 0)\right] + \epsilon},\\
&c_i = \mathrm{physics\text{-}based\ counterfactual}.
\end{aligned}
\end{equation}
where $\delta_{\mathrm{phys},i}$ denotes the magnitude of the
risk-increasing perturbation applied to construct pair $i$.
Pairs with larger perturbations carry clearer directional evidence and
are given proportionally greater influence.

The source-weighted pairwise loss aggregates over all $N$ comparisons:
\begin{equation}
\mathcal{L}_{\mathrm{pair}} =
\frac{1}{N} \sum_{i=1}^{N} w_i \cdot \ell_{\mathrm{BT}}(o_w^{(i)}, o_l^{(i)}).
\end{equation}

\paragraph{Collision-endpoint anchoring.}
Pairwise comparisons shape the relative ordering of $R_\phi$ but leave its absolute scale unconstrained. Collision-imminent states—those occurring in a narrow tolerance window around the moment of impact—provide a natural and consistent anchor: they represent the most dangerous condition observable in the data and carry a uniform interpretation across all events.
We exploit this by imposing two complementary constraints on states drawn from conflict sequences.

Let $I$ denote the set of impact-window states.
We anchor these to a high-risk reference value $r^*$ via a regression penalty:
\begin{equation}
\mathcal{L}_{\mathrm{impact}} =
\frac{1}{|I|}
\sum_{x \in I}
\bigl(R_\phi(x) - r^*\bigr)^2.
\end{equation}
This term ensures that the upper end of the scoring scale is consistently associated with the most dangerous states, providing a stable, data-grounded ceiling.

Let $U$ denote conflict states that lie outside the impact tolerance window—states that are part of a conflict sequence but have not yet reached the critical endpoint.
Although these states are dangerous, they should not exceed the
impact-level reference. We enforce this with an asymmetric upper-bound penalty:
\begin{equation}
\mathcal{L}_{\mathrm{upper}} =
\frac{1}{|U|}
\sum_{x \in U}
\max\!\bigl(R_\phi(x) - u,\;0\bigr)^2,
\end{equation}
where $u < r^*$ is a threshold below the impact reference.
This term prevents earlier conflict states from being scored above the
impact reference, which would invert the expected temporal risk ordering
without being penalised by the pairwise loss.

\paragraph{Overall objective.}
The full training objective combines ordinal learning from comparisons
with scale anchoring from collision endpoints:
\begin{equation}
\mathcal{L} =
\mathcal{L}_{\mathrm{pair}}
+ \lambda_{\mathrm{impact}}\,\mathcal{L}_{\mathrm{impact}}
+ \lambda_{\mathrm{upper}}\,\mathcal{L}_{\mathrm{upper}},
\label{eq:total_loss}
\end{equation}
where $\lambda_{\mathrm{impact}}$ and $\lambda_{\mathrm{upper}}$ balance
the relative contributions of the two regularization terms.
Minimizing $\mathcal{L}$ simultaneously shapes the ordinal structure of
$R_\phi$ through multi-source comparison supervision and anchors its
scale to collision-imminent reference states.

\subsection{Risk-Scoring Function Parameterizations}
\label{sec:instantiations}

We develop three parameterizations of $R_\phi$ that address distinct
practical scenarios: learning a risk score from scratch without any
pre-designed indicators, aligning a single pretrained surrogate model
toward collision-relevant ordering, and combining several complementary
surrogates into a unified aligned score.
The objective~(\ref{eq:total_loss}) is agnostic to this choice. It operates solely on the scalar output of $R_\phi$ and imposes no constraints on its internal structure. It applies uniformly across all three parameterizations.

\subsubsection{Direct Neural Risk-Score Learning (DRL)}
\label{sec:dra}

The first instantiation constructs a risk scoring function entirely from
comparison data, without relying on any pre-designed surrogate indicator.
We parameterise $R_\phi$ as a multilayer perceptron that takes the
13-dimensional vehicle interaction feature vector (Section~\ref{sec:features})
as input and outputs a scalar risk score:
\begin{equation}
R_\phi(o) = f_\phi(o).
\end{equation}
All parameters are optimised end-to-end under the objective~(\ref{eq:total_loss}).
This instantiation imposes no prior structure on the risk function,
allowing it to discover non-linear interactions among input features
that are not captured by any single hand-crafted indicator.

\subsubsection{Single-Surrogate Risk Alignment (SSRA)}
\label{sec:sra}

The second instantiation fine-tunes an existing surrogate model under the
pairwise alignment objective, without modifying its architecture or
introducing additional parameters.
We use GSSM \citep{Jiao2026GSSM} as a representative case. It is pretrained on naturalistic driving data and leverages a distribution-fitting objective over normal driving patterns, which does not directly optimize the collision-relevant ordering in Eq.~(\ref{eq:rank_objective}). This raises the question of whether such pretrained representations can be effectively adapted to collision-relevant risk ordering using only ordinal supervision without retraining or architectural changes.

\paragraph{GSSM risk score.}
GSSM models inter-vehicle spacing $s$ as lognormal given interaction
context $X$, with context-dependent parameters
$\bigl(\hat{\mu}(X), \hat{\sigma}^2(X)\bigr)$ produced by a
context encoder--decoder network.
The CDF of this distribution:
\begin{equation}
F_S\!\left(s;\,\hat{\mu}(X),\hat{\sigma}^2(X)\right)
= \frac{1}{2}\!\left[
    1 - \operatorname{erf}\!\left(
        \frac{\ln s - \hat{\mu}(X)}{\sqrt{2\,\hat{\sigma}^2(X)}}
    \right)
  \right]
\label{eq:gssm_cdf}
\end{equation}
gives the percentile of the current spacing in the lower tail of the
learned normal-interaction distribution.
A high $F_S$ value indicates an unusually tight gap for the given
driving context.

\paragraph{Fine-tuning procedure.}
We use $F_S$ directly as the risk score:
\begin{equation}
R_\phi(o) = F_S\!\left(s;\,\hat{\mu}_\phi(X),\,\hat{\sigma}^2_\phi(X)\right),
\label{eq:sra_score}
\end{equation}
where $\phi$ denotes the full encoder--decoder parameter set.
The objective~(\ref{eq:total_loss}) is applied to this score and
gradients are back-propagated through the CDF into the encoder and
decoder, reshaping the learned spacing distribution so that high-$F_S$
states better reflect the collision-relevant ordering encoded in
comparison data.

\subsubsection{Multi-Surrogate Risk Alignment (MSRA)}
\label{sec:rasc}

The third instantiation addresses the limitation of any single surrogate
by learning a weighted combination of multiple indicators under the
alignment objective. We assume that $V(o)$ is monotone non-decreasing in each surrogate when the others are held fixed. This condition holds by construction for each indicator we consider, ensuring that a positive-weighted combination preserves the correct ordering direction.
The alignment objective then selects weights that best fit the
collision-relevant ordering in $\mathcal{D}$, without requiring any
individual surrogate to be well-calibrated on its own.

\paragraph{Parametric form.}
Since the surrogates operate on heterogeneous numerical scales and some
decrease with risk while others increase, we apply a learnable
affine-sigmoid transformation to each before combining:
\begin{equation}
R_\phi(o) = \sum_{k=1}^{K} w_k \cdot
\sigma\left(\alpha_k \cdot f_k(o) + \beta_k\right),
\label{eq:rasc}
\end{equation}
where $f_k(o)$ is the output of the $k$-th surrogate and
$\{w_k, \alpha_k, \beta_k\}$ are learnable scalars.
The sign of $\alpha_k$ automatically accounts for the surrogate's
risk direction.
All parameters $\phi = \{w_k, \alpha_k, \beta_k\}_{k=1}^K$ are
optimised jointly under~(\ref{eq:total_loss}).

We instantiate this with three surrogates ($K=3$): TAdv, EI, and GSSM.
The learned weights $\{w_k\}$ directly quantify each surrogate's
relative contribution to collision-relevant risk ordering, making the
combined score interpretable in settings where each component carries
established physical meaning.

\section{Experiments}
\label{sec:experiments}

\subsection{Datasets and Data Preparation}
\label{sec:data}

\paragraph{100-Car NDS.}
The 100-Car Naturalistic Driving Study \citep{Dingus2006} recorded
approximately 2 million miles of continuous driving from 100 instrumented vehicles
over a 12-month period. 
We use the reconstructed bird’s-eye-view trajectories of safety-critical events (crashes and near-crashes) of it \citep{jiao2024hundredcarreconstruction}.
Conflict sequences span from $\min(t_{\mathrm{start}},\ t_{\mathrm{impact}} - 4.5)$
to $\min(t_{\mathrm{end}},\ t_{\mathrm{impact}} + 0.5)$ seconds relative to impact.
Safe sequences are drawn from the same reconstructed trajectories in interaction
windows with no flagged event, subject to a minimum 3-second gap from any event,
a minimum duration of 2 seconds, and a maximum absolute deceleration of
$1.5\ \mathrm{m/s^2}$ to exclude latent-hazard frames. After processing, the dataset
yields 180 conflict sequences and 129 safe sequences for training,
validation, and comparison data construction.

\paragraph{SHRP2 NDS.}
The Second Strategic Highway Research Program Naturalistic Driving Study (SHRP2 NDS) \citep{antin2019shrp2methods,hankey2016shrp2database} is a substantially larger naturalistic driving dataset. We use the reconstructed bird’s-eye-view trajectories of safety-critical events (crashes and near-crashes) from \citep{jiao2026shrpcrash}. The SHRP2 crash subset is used exclusively as an out-of-distribution test set; no SHRP2 data are involved in any training or comparison construction steps. Differences in recording equipment, geographic coverage, and participant demographics make SHRP2 a meaningful benchmark for cross-dataset generalisation.

\paragraph{MetaDrive simulation.}
We generate supplementary conflict and safe interaction sequences using MetaDrive simulation \citep{Li2022}. Both the ego vehicle and surrounding traffic agents are controlled by the Intelligent Driver Model (IDM) \citep{Treiber2000IDM}, which governs each vehicle's longitudinal acceleration as
\begin{equation}
\begin{aligned}
&\dot{v} = a\!\left[1 - \left(\frac{v}{v_0}\right)^\delta 
- \left(\frac{s^*(v,\Delta v)}{s}\right)^2\right], \\
&s^*(v,\Delta v) = s_0 + vT + \frac{v\,\Delta v}{2\sqrt{ab}}.
\end{aligned}
\label{eq:idm}
\end{equation}
where $s$ is the current inter-vehicle gap, $\Delta v$ is the relative speed to the leading vehicle, $v_0$ is the desired speed, $T$ is the desired time headway, $s_0$ is the minimum stationary gap, and $a$, $b$ are the maximum acceleration and comfortable deceleration, respectively. The desired gap $s^*$ links the two variables most directly associated with collision risk—inter-vehicle spacing and closing speed—to the vehicle's braking response, ensuring that conflict dynamics generated in simulation are physically grounded in the same feature space as the observation vector $o_t$. Conflicts arise organically from the interplay between IDM-driven agents and road geometry rather than from scripted scenarios. Ten environment configurations are used, combining road topologies (straight segments, curves, X- and T-intersections, roundabouts, and highway ramps), lane counts (2--4), and traffic densities (0.15--0.6), with vehicle positions randomly initialised per scenario. This design produces diverse conflict geometries spanning rear-end collisions on straight and curved roads, angular conflicts at intersections, and merging conflicts at ramps. Trajectories are recorded at 10~Hz over a 15-second pre-impact window and mapped to the same 13-dimensional observation space as the naturalistic data. Simulated data are used only in the comparison source ablation of Section~\ref{sec:ablation}.

\subsection{Observation Features}
\label{sec:features}

We adopt the same observation representation as in GSSM~\citep{jiao2026shrpcrash}:
\begin{equation}
\begin{aligned}
o_t = [&
l_{\mathrm{ego}},\ l_{\mathrm{sur}},\ w_{\mathrm{comb}},\
v_{\mathrm{ego}}^{y},\ v_{\mathrm{sur}}^{x},\ v_{\mathrm{sur}}^{y}, \\
&
(v_{\mathrm{ego}})^2,\ (v_{\mathrm{sur}})^2,\
\Delta v^2,\ \Delta v,\ \psi_{\mathrm{sur}},\ \rho,\ s
].
\end{aligned}
\label{eq:observation}
\end{equation}
where $l$, $w$ are vehicle dimensions; $v^x$, $v^y$ are lateral and longitudinal
velocity components; $\Delta v$ is signed relative speed; $\psi_{\mathrm{sur}}$ is
the partner's heading angle relative to the ego; $\rho$ is the azimuth of the partner
in the relative-velocity frame; and $s$ is the inter-vehicle gap. This representation
is shared across all methods, enabling fair comparison under identical information
conditions.

\subsection{Evaluation Protocol}
\label{sec:eval}

Since frame-level risk probabilities cannot be observed, we assess risk score quality through proactive collision warning. Given a risk threshold $\tau$, the model issues a warning at the first time step $t_{\mathrm{alert}}$ within a sequence at which $R_\phi(o_t) \geq \tau$. For a crash or safety-critical sequence, a successful warning produces a lead time of $t_{\mathrm{end}} - t_{\mathrm{alert}}$, where $t_{\mathrm{end}}$ denotes the impact time for crashes or the safety-critical endpoint for near-crash events. A failure to warn before the endpoint is counted as a miss, while a warning on a safe sequence is counted as a false alarm. Several metrics are reported to evaluate high-sensitivity warning performance, as follows::

\begin{itemize}[leftmargin=*, nosep]

\item The normalized partial area under the receiver operating characteristic (ROC) curve measures the discrimination ability of a risk score in the high-recall region, where missed dangerous events are strictly limited. Let $r$ denote the true positive rate, equivalently recall, and let $\mathrm{FPR}(r)$ denote the false positive rate on the ROC curve at recall level $r$. The normalized partial ROC area is defined as
\begin{equation}
A^{\mathrm{ROC}}_{\alpha}
=
\frac{1}{1-\alpha}
\int_{\alpha}^{1}
\left(1-\mathrm{FPR}(r)\right)\,dr.
\end{equation}
A larger value indicates that the model maintains a lower false-alarm rate while detecting most dangerous events. In this work, $A^{\mathrm{ROC}}_{90\%}$ and $A^{\mathrm{ROC}}_{80\%}$ evaluate discrimination performance when the false-negative rate does not exceed $10\%$ and $20\%$, respectively.

\item Precision in the high-recall region of the precision–recall curve (PRC) measures the reliability of warnings when the system is required to detect most dangerous events. It is defined as the maximum precision achievable on the precision-recall curve under a recall constraint:
\begin{equation}
P^{\mathrm{PRC}}_{\alpha}
=
\max_{\mathrm{Recall}\geq \alpha}
\mathrm{Precision}.
\end{equation}
These metrics focus on whether the model can maintain high warning precision under high-sensitivity operation. In this work, $P^{\mathrm{PRC}}_{90\%}$ and $P^{\mathrm{PRC}}_{80\%}$ correspond to recall requirements of at least $90\%$ and $80\%$, respectively.

\item Precision-constrained mean warning lead time measures how early reliable warnings can be issued. For a threshold $\tau$, let $\mathcal{D}^{+}_{\tau}$ denote the set of correctly warned crash or safety-critical sequences, and let $t^{(i)}_{\mathrm{end}}$ and $t^{(i)}_{\mathrm{alert}}$ denote the endpoint time and alert time of sequence $i$. The metric is defined as
\begin{equation}
\mathrm{mTTI}^{90\%}_p
=
\max_{\tau:\,\mathrm{Precision}(\tau)\geq 0.90}
\frac{1}{|\mathcal{D}^{+}_{\tau}|}
\sum_{i\in\mathcal{D}^{+}_{\tau}}
\left(
t^{(i)}_{\mathrm{end}}
-
t^{(i)}_{\mathrm{alert}}
\right).
\end{equation}
This metric evaluates the maximum mean warning lead time among thresholds that achieve precision of at least $90\%$, thereby measuring warning timeliness under a reliability constraint.

\end{itemize}

Together, these metrics capture three operationally relevant properties of a proactive warning system: high detection coverage of dangerous events, reliable warning precision under high-sensitivity operation, and sufficient warning lead time before impact or the safety-critical endpoint.

\subsection{Baselines and Implementation}
\label{sec:setup}

We compare against four existing methods with four representative surrogate risk scores:
\textbf{TAdv} \citep{Laureshyn2010TAdv}, \textbf{EI} \citep{Cheng2025EI}, and \textbf{TTC2D}
\citep{Guo2023TTC2D} as kinematic SSMs; and \textbf{GSSM} \citep{Jiao2026GSSM} as a distribution-based score, used both as a standalone baseline and as the base model for SSRA.

DRL uses a multilayer perceptron to learn a risk score directly from interaction features, trained end-to-end with the proposed comparison-based ordinal learning objective. SSRA directly fine-tunes GSSM's model by propagating alignment gradients through the CDF into the encoder--decoder without introducing additional parameters. MSRA parameterizes the risk score as an aligned combination of multiple surrogate indicators, using TAdv, EI, and GSSM as the default inputs, and optimizes the combination weights under the same ordinal supervision framework.

All experiments use 5-fold event-level cross-validation on 100-Car NDS, reporting means and standard deviations over multiple random seeds. Each training fold yields approximately 10,000 comparison pairs from temporal, cross-sequence, and physics-based
sources. Out-of-distribution evaluation uses SHRP2 crash events, with aligned models trained on the full 100-Car NDS.

\subsection{Direct Neural Risk-Score Learning}
\label{sec:dra_eval}

A central question for the proposed framework is whether pairwise comparison signals, constructed from naturalistic driving data without any per-frame risk labels, provide a viable direction for driving risk quantification. The DRL setting is particularly suited to investigating this question: by training from scratch without relying on surrogate risk objectives, it isolates the effect of the ordinal supervision provided by the comparison data, enabling a direct evaluation of the effectiveness of the proposed alignment framework.

Tables~\ref{tab:100car} and~\ref{tab:shrp2} present the in-distribution and
out-of-distribution results. 

\begin{table}[ht]
\centering
\scriptsize
\setlength{\tabcolsep}{3pt}
\centering
\caption{Proactive collision warning on 100-Car NDS (in-distribution,
5-fold event-level CV).}
\label{tab:100car}
\resizebox{0.49\textwidth}{!}{%
\begin{tabular}{lccccc}
\toprule
Method &
$A^{\mathrm{ROC}}_{90\%}$ (\%) &
$A^{\mathrm{ROC}}_{80\%}$ (\%) &
$P^{\mathrm{PRC}}_{90\%}$ (\%) &
$P^{\mathrm{PRC}}_{80\%}$ (\%) &
$\mathrm{mTTI}^{90\%}_p$ (s) \\
\midrule
TTC2D & 80.11 & 86.75 & 65.41 & 99.33 & 3.29 \\
TAdv  & 83.97 & 89.15 & 98.09 & 98.66 & 2.94 \\
EI    & 83.86 & 90.26 & 97.26 & 99.51 & 3.45 \\
GSSM  & 82.83 & 88.29 & 97.16 & 98.88 & 3.49 \\
\midrule
DRL(Ours)                & \textbf{97.06} & \textbf{97.58} & \textbf{99.43} & \textbf{99.55} & \textbf{4.32} \\
\bottomrule
\end{tabular}
}
\end{table}

\begin{table}[ht]
\centering
\scriptsize
\setlength{\tabcolsep}{3pt}
\caption{Proactive collision warning on SHRP2 NDS crash events
(out-of-distribution; aligned models trained on full 100-Car NDS).}
\label{tab:shrp2}
\resizebox{0.49\textwidth}{!}{%
\begin{tabular}{lccccc}
\toprule
Method &
$A^{\mathrm{ROC}}_{90\%}$ (\%) &
$A^{\mathrm{ROC}}_{80\%}$ (\%) &
$P^{\mathrm{PRC}}_{90\%}$ (\%) &
$P^{\mathrm{PRC}}_{80\%}$ (\%) &
$\mathrm{mTTI}^{90\%}_p$ (s) \\
\midrule
TTC2D & 52.00 & 52.00 &  0.00 &  0.00 & 1.83 \\
TAdv  & 44.20 & 62.46 & 75.60 & 86.89 & 0.75 \\
EI    & 45.00 & 57.46 & 70.20 & 92.86 & 2.09 \\
GSSM  & 76.50 & 85.62 & \textbf{93.70} & \textbf{96.43} & 2.63 \\
\midrule
DRL(Ours)                 & \textbf{85.85} & \textbf{89.85} & 93.65 & 96.30 & \textbf{5.96} \\
\bottomrule
\end{tabular}}
\end{table}

On 100-Car NDS, DRL achieves the strongest discrimination ($A^{\mathrm{ROC}}_{90\%} =
97.06\%$) and the longest warning lead time ($\mathrm{mTTI}^{90\%}_p = 4.32$ s), substantially exceeding all unaligned baselines. The out-of-distribution results are particularly informative. On SHRP2, kinematic SSMs degrade substantially under dataset shift: TAdv's lead time collapses to $0.75$ s, and EI and TTC2D perform similarly. DRL, by contrast, maintains $\mathrm{mTTI}^{90\%}_p = 5.96$ s, the longest warning lead time across both datasets. These results confirm that the ordinal supervision encoded in the comparison data provides a robust training signal: a model guided solely by collision-anchored pairwise constraints generalises across recording conditions and detects conflicts earlier than kinematic proxies calibrated against surrogate objectives.

\begin{figure*}
\centering

\begin{subfigure}{0.8\linewidth}
    \centering
    \includegraphics[width=\linewidth]{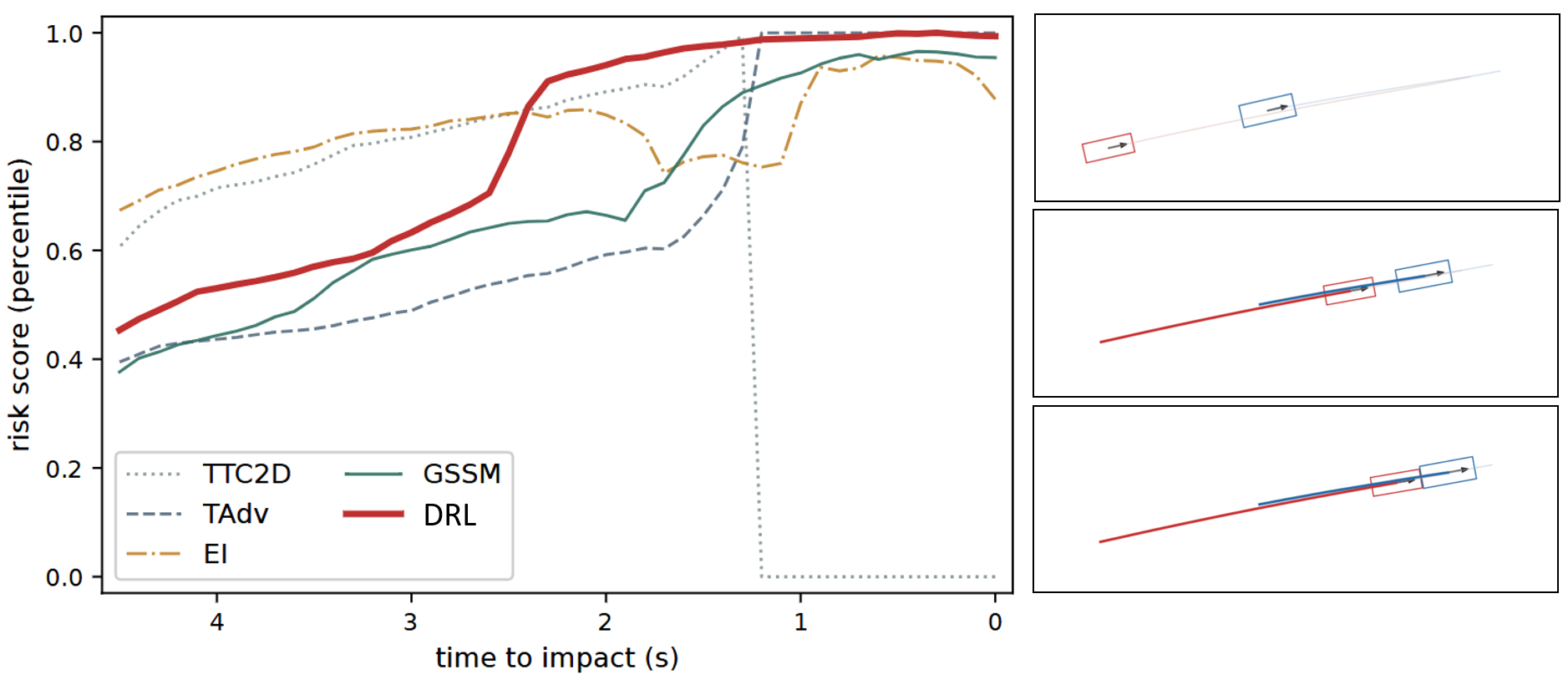}
    \caption{Case 1}
\end{subfigure}
\begin{subfigure}{0.8\linewidth}
    \centering
    \includegraphics[width=\linewidth]{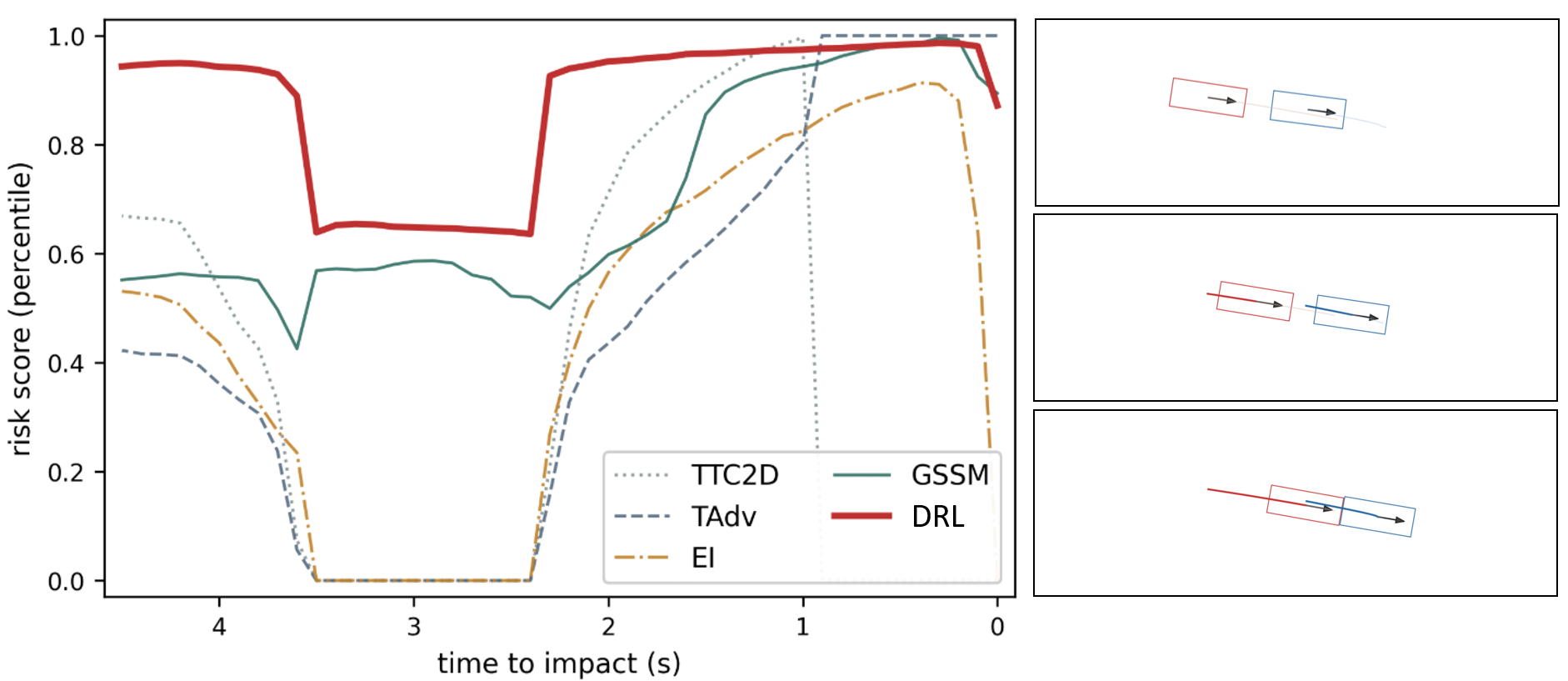}
    \caption{Case 2}
\end{subfigure}

\begin{subfigure}{0.8\linewidth}
    \centering
    \includegraphics[width=\linewidth]{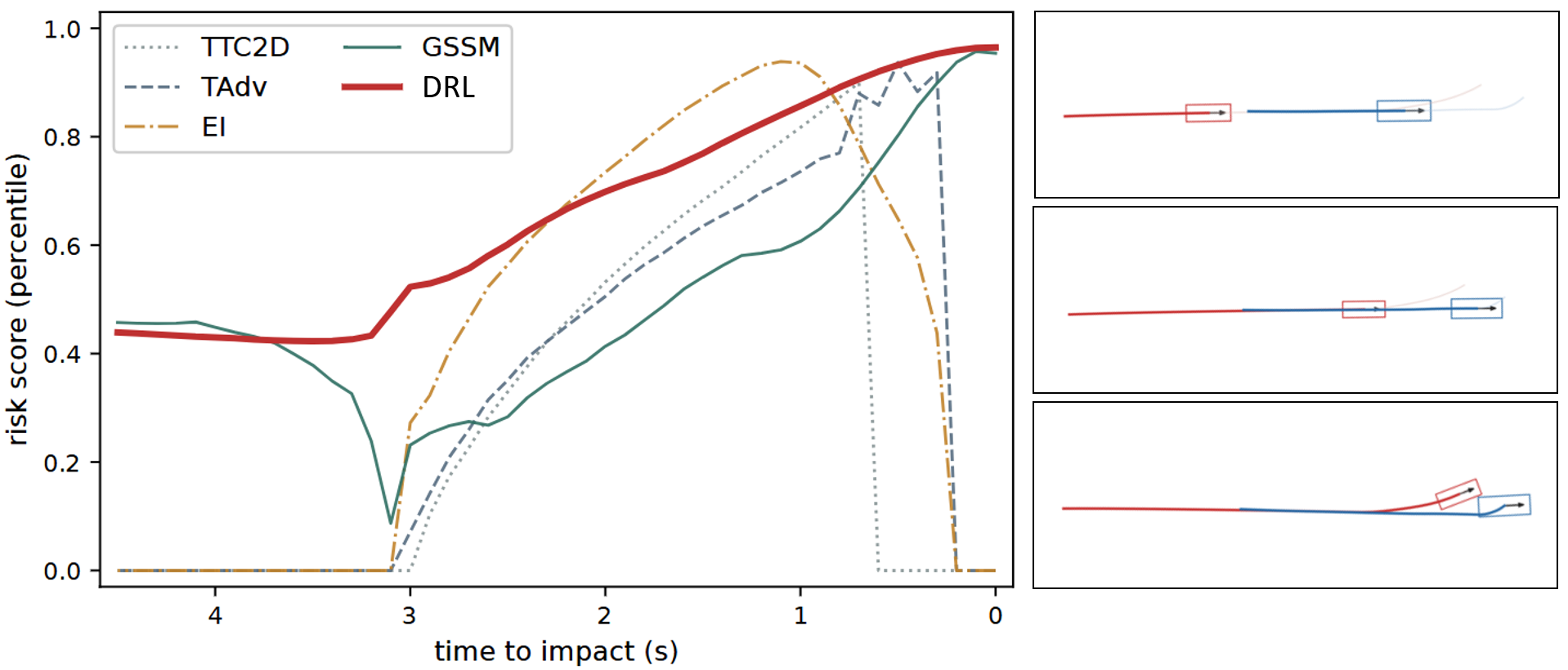}
    \caption{Case 3}
\end{subfigure}
\caption{Risk score trajectories in three representative scenarios.
}
\label{fig:three_cases}
\end{figure*}

Figure~\ref{fig:three_cases} illustrates the qualitative behaviour of the learned risk scores on three representative 100-Car crash events. In Cases 1 and 3, DRL produces smooth and temporally consistent trajectories that increase progressively as impact approaches, suggesting that the learned score captures the accumulation of collision risk. Case 2 further shows that DRL is not merely learning a monotonic time-to-impact trend. Its score remains elevated throughout most of the hazardous interaction and avoids the near-zero collapse observed in several kinematic surrogates, while still exhibiting a moderate mid-sequence decrease when the relative speed between the vehicles temporarily becomes smaller. The score then rises again as the conflict re-intensifies toward impact. This behaviour indicates that DRL preserves persistent conflict-level danger while remaining responsive to transient changes in vehicle dynamics. In contrast, the unaligned baselines often exhibit delayed responses, abrupt saturation or drops, and local non-monotonic fluctuations.

\subsection{Single-Surrogate Risk Alignment}
\label{sec:sra_eval}

Beyond learning a risk scoring function from scratch, the framework is equally applicable to pre-trained surrogate models. A key practical advantage is that alignment can be applied without modifying a model's architecture or retraining it from scratch: the Bradley--Terry loss can be propagated through an existing model's parameters, redirecting its output toward collision-relevant ordering while preserving the structure that was established during pre-training. We demonstrate this using GSSM as the base model. GSSM's parameters are fine-tuned directly under the Bradley--Terry loss through the existing CDF computation, with no new components introduced. Temporal comparison pairs from 100-Car NDS serve as the alignment signal.

\begin{table}[ht]
\centering
\scriptsize
\setlength{\tabcolsep}{3pt}
\caption{SSRA applied to GSSM on 100-Car NDS
(5-fold event-level CV).}
\label{tab:sra_gssm_posthoc}
\resizebox{0.49\textwidth}{!}{%
\begin{tabular}{lccccc}
\toprule
Method & $A^{\mathrm{ROC}}_{90\%}$ (\%) & $A^{\mathrm{ROC}}_{80\%}$ (\%) & $P^{\mathrm{PRC}}_{90\%}$ (\%) & $P^{\mathrm{PRC}}_{80\%}$ (\%) & $\mathrm{mTTI}^{90\%}_p$ (s) \\
\midrule
GSSM & 82.82 & 88.29 & 97.16 & 98.88 & 3.49 \\
SSRA   & \textbf{83.18} & \textbf{89.35} & \textbf{97.35} & \textbf{98.92} & \textbf{3.54} \\
\bottomrule
\end{tabular}}
\end{table}

As shown in Table~\ref{tab:sra_gssm_posthoc}, SSRA yields consistent improvements across all metrics relative to the unaligned GSSM baseline. The gains are modest, which reflects that GSSM's distributional objective already captures aspects of collision-relevant risk structure, and that alignment operating through the existing scalar output path is naturally bounded by what that path preserves. The consistent positive direction across metrics nonetheless demonstrates that the alignment framework is compatible with existing surrogate-based parameterized risk models and can improve their collision-predictive ordering without architectural redesign—a useful property when deploying or updating models in operational settings.

\subsection{Multi-Surrogate Risk Alignment}
\label{sec:rasc_eval}

Where DRL and SSRA each address single-model scenarios, MSRA addresses the common situation in which multiple established surrogate indicators are available.
Each indicator captures a genuine dimension of collision risk but operates on a different numerical scale and with different failure modes; no principled basis for combining them has previously been established. MSRA fills this gap by learning combination weights under the pairwise alignment objective, so that each indicator contributes according to its collision-predictive value rather than by engineering convention.

On 100-Car NDS, MSRA achieves the highest precision among the evaluated methods. The TAdv+EI+GSSM configuration obtains $A^{\mathrm{ROC}}_{90\%} = 96.46\%$ and $P^{\mathrm{PRC}}_{90\%} = 100.00\%$, while the EI+GSSM configuration attains a slightly longer $mTTI^{90}_{p}$ of 3.66~s. Its warning lead time remains below that of DRL, indicating that the expressivity of the aligned combination is still bounded by the constituent surrogates. The complementarity between MSRA and DRL is therefore reflected in the 100-Car evaluation: MSRA provides an interpretable high-precision surrogate fusion, whereas DRL achieves the longest warning lead time.

The joint results in Table~\ref{tab:msra_ablation_weights} provide both the warning performance of each MSRA
configuration and the corresponding learned weight assignment, offering an interpretable account of how the alignment
objective uses each indicator. EI and GSSM consistently receive positive weights, while TAdv receives a negative weight when included. This pattern is semantically coherent: EI and GSSM are defined such that higher values indicate greater risk, whereas TAdv is defined such that lower values indicate reduced temporal margin before potential conflict. The learned weight signs directly recover these directional relationships, assigning each indicator a contribution consistent with its established engineering interpretation. The alignment objective thus not only improves overall warning performance but also
produces a combination that is semantically well-grounded in the constituent indicators.

\begin{table*}[t]
\centering
\scriptsize
\setlength{\tabcolsep}{6pt}
\caption{MSRA indicator ablation on 100-Car NDS and the corresponding learned weights after sigmoid normalisation.}
\label{tab:msra_ablation_weights}
\footnotesize
\resizebox{0.90\textwidth}{!}{%
\begin{tabular}{lcccccccc}
\toprule
& \multicolumn{5}{c}{Performance} & \multicolumn{3}{c}{Weight assignment} \\
\cmidrule(lr){2-6} \cmidrule(lr){7-9}
Configuration 
& $A^{\mathrm{ROC}}_{90\%}$ (\%) 
& $A^{\mathrm{ROC}}_{80\%}$ (\%) 
& $P^{\mathrm{PRC}}_{90\%}$ (\%) 
& $P^{\mathrm{PRC}}_{80\%}$ (\%) 
& $\mathrm{mTTI}^{90}_{p}$ (s) 
& TAdv 
& EI 
& GSSM \\
\midrule
EI+GSSM 
& 96.08 & 98.04 & 100.00 & 100.00 & 3.66 
& -- & 1.714 & 1.810 \\
TAdv+EI+GSSM 
& 96.46 & 98.23 & 100.00 & 100.00 & 3.58 
& $-0.607$ & 1.503 & 1.649 \\
\bottomrule
\end{tabular}}
\end{table*}

\subsection{Analysis of Pairwise Supervision Sources}
\label{sec:ablation}

\subsubsection{Individual Source Informativeness and Data Scaling}

The proposed learning framework draws pairwise supervision from three complementary real-data comparison types: temporal comparisons, cross-sequence comparisons, and physics-based counterfactual comparisons. In addition, simulation-derived comparison pairs are evaluated as an auxiliary augmentation source rather than as an independent comparison rule. To characterize the contribution of each source individually and in combination, we train DRL on each source independently and on all three combined, varying the fraction of 100-Car training events from 10\% to 60\%.
This design allows both the informativeness of each source and the benefit of combining multiple sources to be assessed under comparable conditions.

\begin{figure*}
\centering
\includegraphics[width=0.9\linewidth]{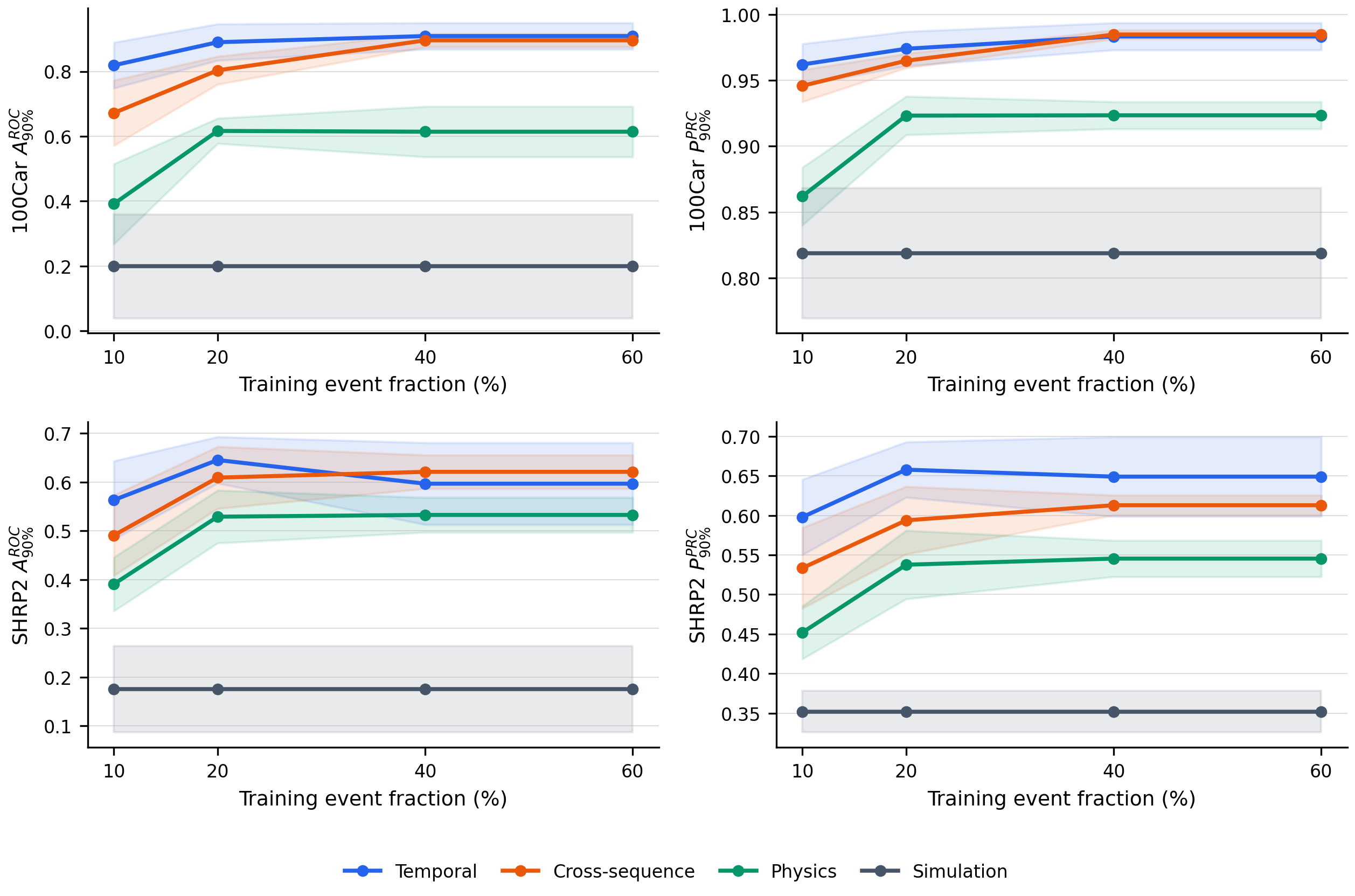}
\caption{Individual informativeness of pairwise supervision sources for DRL under different training event fractions.}
\label{fig:dra_scaling}
\end{figure*}

Figure~\ref{fig:dra_scaling} reports DRL performance when each supervision source is used independently under different training event fractions. Temporal and cross-sequence comparisons provide the strongest standalone supervision and improve consistently with the number of training events, reaching $A^{\mathrm{ROC}}_{90\%}=0.909$ and $0.896$, respectively, at the $60\%$ scale. Physics-based counterfactual comparisons improve rapidly at low data volumes but plateau at approximately $0.614$. This trend suggests that their local monotonic constraints are useful for establishing the directional effects of risk-relevant variables, but are insufficient by themselves to capture full event-level risk progression. Simulation-derived comparisons perform poorly when used alone, indicating that simulated interactions cannot replace real safety-critical event data because of the distribution gap from naturalistic driving.

\begin{figure*}
\centering
\includegraphics[width=0.9\linewidth]{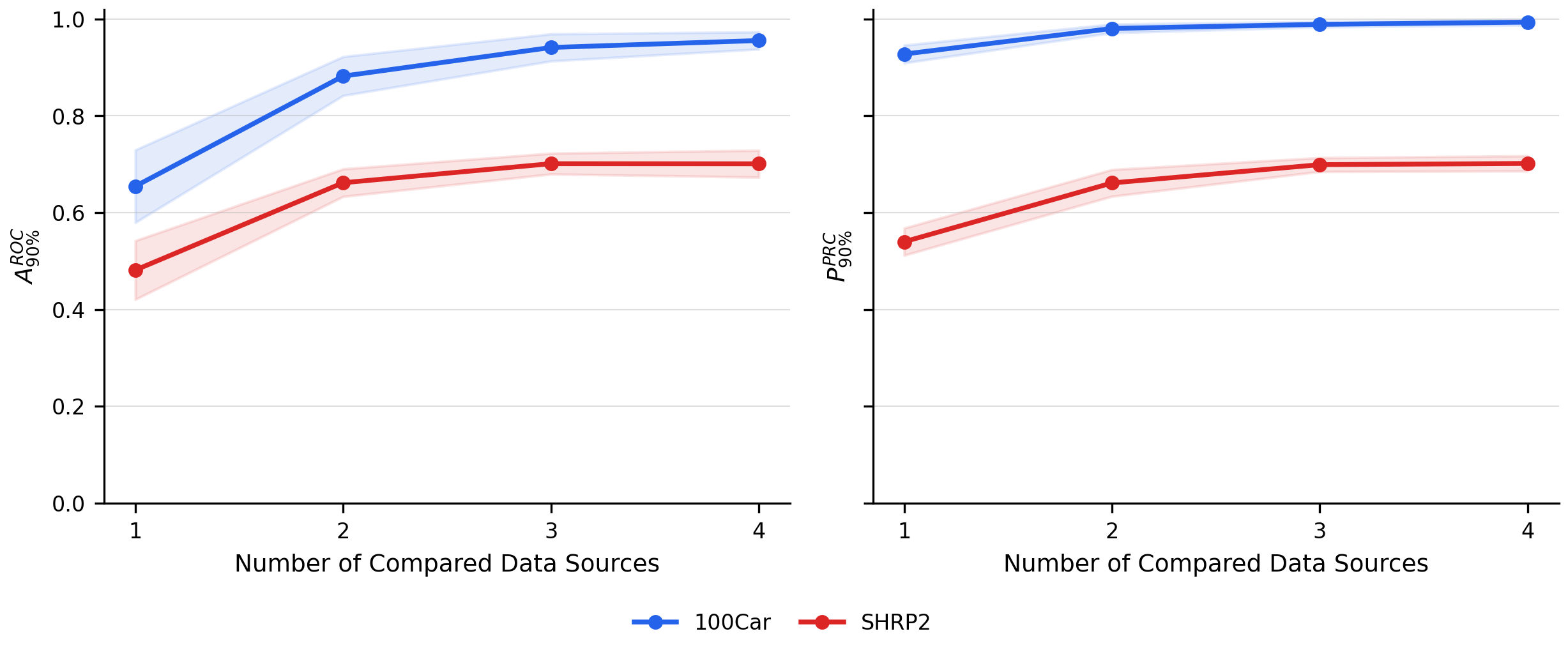}
\caption{DRL performance aggregated by subset size over all combinations of three real-data comparison sources and one simulation-derived augmentation source.}
\label{fig4}
\end{figure*}

\begin{figure*}
\centering
\includegraphics[width=0.9\linewidth]{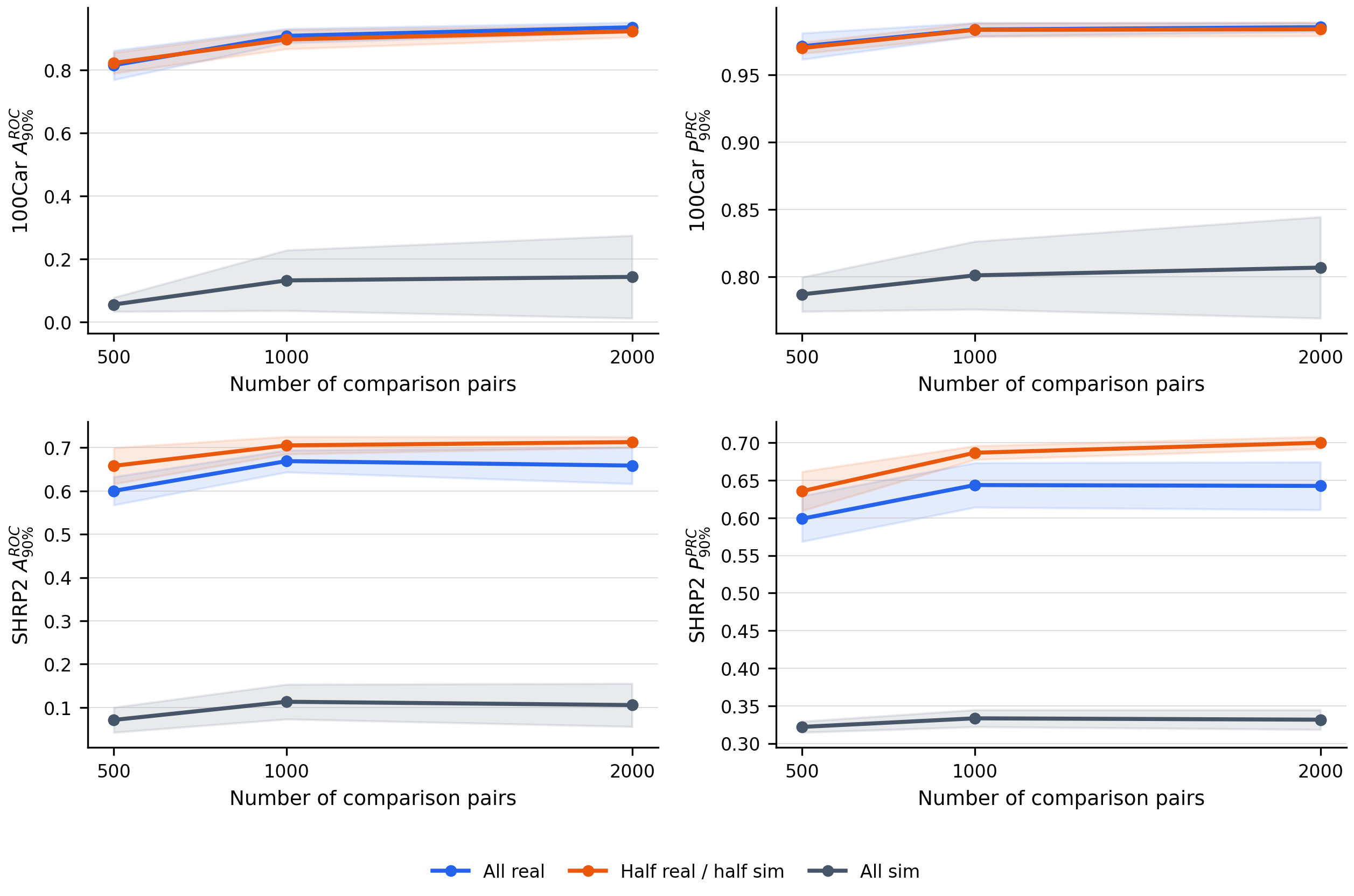}
\caption{Effect of incrementally adding MetaDrive simulation comparison pairs on DRL performance, evaluated at small and large real-data budgets.}
\label{fig:sim}
\end{figure*}

\subsubsection{Simulation as a Supplementary Comparison Source}
\label{sec:sim_ablation}

To assess source complementarity, we further train DRL on all source subsets of size $k$, formed from the three real-data comparison sources and one simulation-derived augmentation source, and aggregate the results by subset size. As shown in Figure~\ref{fig4}, $A^{\mathrm{ROC}}_{90\%}$ increases from $0.655$ at $k=1$ to $0.956$ at $k=4$, showing that additional supervision sources provide incremental benefit. The improvement is most pronounced when moving from a single source to multiple sources, while the marginal gain becomes smaller as $k$ increases. This pattern suggests that the sources provide complementary ordinal constraints, but also share partial overlap in the risk-ordering information they encode.

Simulation augmentation offers a route to extending the comparison pool when naturalistic conflict data are scarce. We train DRL model on MetaDrive simulation data alone, on real dataset alone, and on combinations of them.

Simulation-only training yields near-chance performance ($A^{\mathrm{ROC}}_{90\%}
\approx 0.20$), confirming that simulated comparisons alone do not transfer to naturalistic driving conditions. When simulation comparisons are added to a small real-data budget, however, measurable gains are observed: $A^{\mathrm{ROC}}_{90\%}$
increases from $0.708$ to approximately $0.835$ as simulation events are incrementally included.
At larger real-data scales, the marginal benefit of simulation is negligible (Figure~\ref{fig:sim}).

These results indicate that simulation can partially compensate for limited naturalistic conflict data in low-data regimes—a practically relevant setting for rare crash types or under-instrumented environments—but cannot substitute for real driving data as the latter accumulates.

\section{Conclusion}
\label{sec:conclusion}

This paper proposed a comparison-based ordinal learning framework for proactive driving risk assessment. The key observation is that naturalistic driving data, despite the scarcity of collision events, contain rich ordinal information. Temporal progression within safety-critical sequences, event-level contrasts between dangerous and normal interactions, and physics-based counterfactual perturbations all indicate which situations should be considered more dangerous, without requiring calibrated frame-level risk labels. By translating these signals into pairwise ordinal supervision, the proposed framework learns collision-relevant risk-score orderings that are easier to obtain than calibrated probabilistic risk labels, while remaining sufficient for the operational objectives of proactive collision warning. Three instantiations were developed to cover complementary risk-scoring parameterizations. DRL demonstrated that event-structured ordinal supervision can serve as an effective training signal for learning risk scores from scratch, achieving strong discrimination and long warning lead time across both in-distribution and out-of-distribution evaluations. SSRA showed that the same ordinal objective can refine an existing surrogate model toward collision-relevant ordering without architectural redesign, yielding consistent improvements over the original GSSM baseline. MSRA demonstrated that comparison-based ordinal learning provides a principled basis for integrating multiple surrogate indicators, with learned weights that preserve the directional semantics of constituent measures and achieve the highest warning precision among the evaluated methods on 100-Car NDS. The source analysis further showed that temporal, cross-sequence, and physics-based counterfactual comparisons provide complementary supervision. Their combination consistently outperformed individual sources, with the benefit most pronounced under limited real-data conditions. Simulation-derived comparisons can partially compensate for scarce naturalistic safety-critical data in low-data regimes, but provide limited marginal gain once sufficient real data are available, confirming their role as auxiliary augmentation rather than a substitute for real event-structured supervision.

Several directions remain open. Extending the framework to more complex multi-agent scenarios and heterogeneous interaction geometries is a natural step toward broader deployment. In such settings, risk-scoring functions should accommodate a variable number of surrounding vehicles and more diverse interaction topologies. The proposed ordinal risk learning framework can be further integrated with sensor-based inputs rather than only kinematic features, which would extend the framework to a wider range of modalities in advance proactive safety applications for intelligent transportation and autonomous driving systems.









\bibliographystyle{elsarticle-harv}
\bibliography{cas-refs}






\end{document}